\documentclass[10pt,twocolumn,letterpaper]{article}

\usepackage{graphicx}
\usepackage{subcaption}
\usepackage{calc}
\usepackage{amssymb}
\usepackage{amstext}
\usepackage{amsmath}
\usepackage{amsthm}
\usepackage{pslatex}
\usepackage{algorithm2e}
\usepackage{tabularx}
\usepackage{booktabs}
\usepackage{rotating}
\usepackage{multirow}
\usepackage[bottom]{footmisc}
\usepackage{tikz}
\usepackage{quantikz}
\usepackage{physics}
\usepackage{cvpr}
\usepackage[pagebackref,breaklinks,colorlinks]{hyperref}

\title{Reinforcement Learning for Parameterized Quantum State Preparation: A Comparative Study}

\author{
\begin{tabular}{cccc}
Gerhard Stenzel & Isabella Debelic & Michael Kölle & Tobias Rohe \\
\end{tabular}
\\
\begin{tabular}{ccc}
Leo Sünkel & Julian Hager & Claudia Linnhoff-Popien \\
\end{tabular}
\\[0.4em]
LMU Munich\\
Department of Computer Science\\
Chair of Mobile and Distributed Systems\\
{\tt\small gerhard.stenzel@ifi.lmu.de}
}

\begin{document}
\maketitle

\begin{abstract}
We extend directed quantum circuit synthesis (DQCS) with reinforcement learning from purely discrete gate selection to parameterized quantum state preparation with continuous single-qubit rotations \(R_x\), \(R_y\), and \(R_z\). We compare two training regimes: a one-stage agent that jointly selects the gate type, the affected qubit(s), and the rotation angle; and a two-stage variant that first proposes a discrete circuit and subsequently optimizes the rotation angles with Adam using parameter-shift gradients. Using Gymnasium and PennyLane, we evaluate Proximal Policy Optimization (PPO) and Advantage Actor--Critic (A2C) on systems comprising two to ten qubits and on targets of increasing complexity with \(\lambda\) ranging from one to five. Whereas A2C does not learn effective policies in this setting, PPO succeeds under stable hyperparameters (one-stage: learning rate approximately \(5\times10^{-4}\) with a self-fidelity-error threshold of 0.01; two-stage: learning rate approximately \(10^{-4}\)). Both approaches reliably reconstruct computational basis states (between 83\% and 99\% success) and Bell states (between 61\% and 77\% success). However, scalability saturates for \(\lambda\) of approximately three to four and does not extend to ten-qubit targets even at \(\lambda=2\). The two-stage method offers only marginal accuracy gains while requiring around three times the runtime. For practicality under a fixed compute budget, we therefore recommend the one-stage PPO policy, provide explicit synthesized circuits, and contrast with a classical variational baseline to outline avenues for improved scalability.
\end{abstract}

\section{Introduction}
\label{sec:introduction}

Directed quantum circuit synthesis (DQCS) formulates the design of quantum circuits as a goal-driven optimization problem: given a target transformation or state, construct a circuit from a fixed gate set that achieves the target with minimal depth or cost. Recent work demonstrates that reinforcement learning (RL) can learn effective editing policies for discrete circuit optimization and for discrete gate selection in DQCS-style environments \cite{rw1,tom}. However, most existing RL formulations treat gate choices as purely discrete and avoid the central difficulty of realistic quantum control, namely the presence of continuous rotation parameters.

This paper extends DQCS to parameterized quantum state preparation with continuous rotation gates (\(R_x, R_y, R_z\)). We study two training strategies: (i) a one-stage agent that selects the gate, the qubit(s), and the angle jointly; and (ii) a two-stage approach in which the agent proposes a discrete circuit topology that is subsequently refined by Adam using parameter-shift gradients \cite{adam,shift}. Our environment is implemented with Gymnasium and PennyLane, and we evaluate Proximal Policy Optimization (PPO) and Advantage Actor--Critic (A2C) on targets spanning two to ten qubits and depth budgets with \(\lambda\in\{1,\dots,5\}\).

We report three main findings. First, A2C fails to learn useful policies in this setting, whereas PPO succeeds with stable hyperparameters. Second, the two-stage strategy provides at most marginal improvements while incurring around three times the runtime, which makes the one-stage approach more practical for parameterized state preparation. Third, both approaches exhibit clear scalability limits: performance saturates for comparatively simple circuits (\(\lambda\leq 3\!\text{--}\!4\)) and degrades sharply for ten-qubit targets even at \(\lambda=2\).

Contributions: (1) a reinforcement learning environment for parameterized state preparation that bridges discrete DQCS and continuous control; (2) a systematic comparison of one-stage and two-stage training for parameterized gate settings; (3) an explicit formalization of the Hilbert-space setting, gate universe, and hybrid Gymnasium--PennyLane workflow, including a worked circuit example; and (4) an empirical characterization of scalability and hyperparameter regimes together with a classical variational baseline for context (for example, PPO with \(\alpha\!\approx\!5\times 10^{-4}\) and a self-fidelity-error threshold of 0.01 for the one-stage regime). In contrast to prior discrete-only DQCS or curriculum-driven architecture search \cite{rw1,rw2}, our results quantify the practical trade-offs that arise when moving to continuous parameters and highlight where future techniques (for example, curriculum design or improved initialization) may be needed. Unlike Fösel et al. \cite{rw1}, who optimize existing circuits via discrete edit operations, we synthesize parameterized state-preparation circuits end-to-end with continuous angles, changing both the action space and the learning dynamics. The remainder of this paper introduces background and related work, details the problem formulation and methodology, presents experimental results, and concludes with a discussion of implications and future directions.

\section{Background and Related Work}
\label{sec:background}

This section summarizes the elements of reinforcement learning and quantum computing relevant to our formulation and positions the work within prior art.

\subsection{Hilbert-Space Model and Parameterized States}
We work on the Hilbert space \(\mathcal{H}_n = (\mathbb{C}^2)^{\otimes n}\) with computational basis \(\{\lvert b\rangle : b\in\{0,1\}^n\}\). A parameterized quantum state is written as
\begin{equation}
    \lvert\psi(\boldsymbol{\theta})\rangle = U(\boldsymbol{\theta}) \lvert 0\rangle^{\otimes n}, \quad U(\boldsymbol{\theta})=\prod_{k=1}^{\ell} U_k(\theta_k),
\end{equation}
where \(\theta_k\in[-\pi,\pi]\) are continuous rotation parameters and \(U_k\in\{R_x,R_y,R_z,\mathrm{CNOT}\}\). Measurement is performed on the statevector \(\lvert\psi\rangle\); when density matrices are required we set \(\rho(\boldsymbol{\theta})=\lvert\psi(\boldsymbol{\theta})\rangle\langle\psi(\boldsymbol{\theta})\rvert\). The fidelity with a target \(\lvert\psi_\mathrm{t}\rangle\) is \(F(\boldsymbol{\theta})=\lvert\langle\psi_\mathrm{t}\mid\psi(\boldsymbol{\theta})\rangle\rvert^2\), which provides the basis for the reward and the self-fidelity error (SFE) \(1-F\).

\subsection{Gate Universe and Parameter Selection}
The gate universe is \(\mathcal{G}=\{R_x(\theta), R_y(\theta), R_z(\theta), \mathrm{CNOT}\}\). Angles \(\theta\) are continuous and drawn from \([-\pi,\pi]\), while \(\mathrm{CNOT}\) uses discrete control--target pairs. When a Clifford+\(T\) fallback is required, we use \(\{H, S, T, \mathbb{1}, \mathrm{CNOT}\}\), but all reinforcement-learning experiments use the rotation-plus-CNOT universe. This gate set is universal and matches the primitives available in PennyLane's \texttt{default.qubit}, which ensures that any synthesized circuit can be executed without further decomposition. Defining the gate universe explicitly clarifies which operations the policy must master and how the action space relates to feasible hardware implementations.

\subsection{Problem Setting}
We formulate parameterized quantum state preparation as a sequential decision process. The agent observes a compact, real-valued encoding of both the current and target state amplitudes and selects, at each step, a gate from \(\{R_x, R_y, R_z, \mathrm{CNOT}\}\), the associated qubit indices, and a rotation angle \(\theta\in[-\pi,\pi]\). Episodes terminate when a self-fidelity error (SFE) threshold is met or a depth budget \(L=2\lambda\) is exhausted. The reward signal is aligned with fidelity so that improvements in \(F=|\langle\psi_t|\psi\rangle|^2\) translate into positive returns. For foundational concepts on qubits, gates, and circuits we refer to the standard text by Nielsen and Chuang \cite{QcandQi}. Among on-policy methods, we experimentally evaluate Advantage Actor--Critic (A2C) and Proximal Policy Optimization (PPO), with PPO typically offering greater stability for mixed discrete--continuous parameterizations.

\subsection{Related Work}
Fösel et al. study deep reinforcement learning (RL) optimization of existing discrete circuits via edit policies defined over a discrete transformation space, without addressing continuous parameters \cite{rw1}. Kölle et al. introduce a DQCS environment in which an agent synthesizes target circuits by selecting discrete gates; parameters remain discrete or implicitly fixed \cite{tom}. Beyond DQCS, curriculum-based RL for quantum architecture search improves robustness under hardware noise, yet operates predominantly in discrete action spaces \cite{rw2}. Other RL approaches target the search for variational quantum circuit architectures \cite{rw3}. Recent quantum machine learning applications also show how hybrid quantum models interface with classical objectives, for example in sequence modeling with selective state space dynamics \cite{qml2}. Efficient statevector simulation techniques such as gate-matrix caching and circuit splitting further support large-scale experimentation on classical backends \cite{qml5}. Our work complements these threads by addressing continuous rotation parameters explicitly and by comparing a one-stage joint selection strategy against a two-stage approach that defers parameter refinement to an optimizer (Adam with parameter-shift gradients) \cite{adam,shift}. For continuous-control stability we adopt PPO \cite{ppo}.

\subsection{Gradient Methods and Parameterized-Action RL}
Classical gradient-based approaches to quantum control and variational algorithms optimize continuous circuit parameters directly using analytic or stochastic gradient estimates. In the variational setting, parameter-shift rules enable unbiased gradient evaluation on quantum circuits, which in turn allows the use of first-order optimizers such as Adam \cite{shift,shiftorigin,adam,cerezo2021vqa}. Stochastic approximation schemes like SPSA provide alternative gradient estimates when analytic rules are impractical or noisy \cite{spall1992spsa}. Our two-stage baseline mirrors this toolbox by refining angles on a fixed discrete scaffold with analytic gradients.

In reinforcement learning, parameterized-action formulations treat actions as a hybrid of discrete choices with associated continuous parameters. Policy-gradient methods naturally extend to such settings by coupling categorical distributions over discrete decisions with continuous distributions over parameters \cite{ppo}. Representative algorithms include early Q-learning for Parameterized-Action Markov Decision Processes (Q-PAMDP) style approaches and deep methods that operate directly in parameterized action spaces \cite{masson2016parameterized,hausknecht2016parameterized}. Our one-stage agent instantiates this paradigm by jointly sampling gate type, qubit indices, and a rotation angle from a mixed policy, while our two-stage baseline separates the discrete selection from continuous refinement.

\subsection{Positioning}
The key novelty is a reinforcement-learning formulation of parameterized state preparation that integrates continuous angles into the DQCS loop together with a controlled comparison of one- versus two-stage training regimes. We report stable hyperparameter regimes and identify a practical scalability boundary (\(\lambda\) of approximately three to four) under PPO, thereby providing guidance for subsequent methods aimed at overcoming these limits.

\section{Problem Formulation and Environment Design}
\label{sec:problem-environment}

\subsection{Objective and Metrics}
Given an \(n\)-qubit target state \(\lvert\psi_\mathrm{t}\rangle\) and a gate set \(\mathcal{G}=\{R_z, \mathrm{CNOT}, R_x, R_y\}\) (with a Clifford+\(T\) fallback \(\{ \mathrm{CNOT}, H, T, S, \mathbb{1}\}\) used only when explicitly selected), we seek a circuit \(\mathcal{C}\in\mathcal{G}^{\le L}\) that prepares \(\lvert\psi_\mathrm{t}\rangle\) from \(\lvert 0\rangle^{\otimes n}\) under a depth budget \(L=2\,\lambda\). The quality of a candidate state \(\lvert\psi\rangle=\mathcal{U}(\mathcal{C})\lvert 0\rangle^{\otimes n}\) is measured by fidelity \(F=\lvert\langle\psi_\mathrm{t}\mid\psi\rangle\rvert^2\). We also track reconstructed circuit depth (RCD; reconstructed depth as a percentage of the target depth), which expresses how efficiently the synthesized circuit realizes the target transformation.

\subsection{State Representation}
We encode both current and target amplitudes as real-valued vectors by concatenating real and imaginary parts: \(x \in \mathbb{R}^{4\cdot 2^n}\). This compact, differentiable representation allows the policy to track proximity to the target without reconstructing full circuits. Targets are generated procedurally by applying exactly \(L/2\) gates sampled from \(\mathcal{G}\) to \(\lvert 0\rangle^{\otimes n}\); intermediate candidates that overlap with previously visited states within a tolerance of 0.1 are rejected to avoid trivial repeats. This procedure fixes difficulty via \(\lambda\) and keeps target generation reproducible.

\subsection{Action Space}
Each action is a tuple \(a=(g,\,q,\,\theta)\), where \(g\in\mathcal{G}\), \(q\) indexes the affected qubit(s) (single-qubit for rotations, control/target pair for CNOT), and \(\theta\in[-\pi,\pi]\) for rotations. We use a mixed parameterization with discrete indices for \(g\) and \(q\) and a continuous angle \(\theta\). For numerical stability, we sample an unconstrained angle \(\tilde{\theta}\in[0,1]\) and map it to the desired interval via \(\theta=2\pi\tilde{\theta}-\pi\).

\subsection{Episode Control and Reward}
Episodes terminate when either the self-fidelity error (SFE) threshold is met, \(1-F \le \varepsilon\), or the depth budget is consumed (\(\lvert\mathcal{C}\rvert=L\)). We set \(\varepsilon=0.01\) unless noted. Circuit execution uses PennyLane's statevector simulator \texttt{default.qubit} without shot noise; the implied measurement operator is the identity on \(\mathcal{H}_n\) (equivalently full-state tomography), and rewards depend solely on fidelity \(F=\lvert\langle\psi_\mathrm{t}\mid\psi\rangle\rvert^2\) rather than on projective or expectation-value measurements. The depth budget is governed by the complexity parameter \(\lambda\) via \(L=2\lambda\). We align the objective with fidelity by using a shaped reward \(r_t=\Delta F_t=F_{t}-F_{t-1}\) with a terminal bonus when \(1-F\le \varepsilon\); equivalently, one can define a cost \(C=1-F\) and reward progress as \(-\Delta C\). To improve stability, we clip extreme values and normalize rewards across minibatches.

\subsection{Algorithms}
We evaluate A2C and PPO as canonical on-policy baselines for continuous control, with PPO ultimately preferred for stability under mixed action parameterizations.

\section{Methodology and Experimental Setup}
\label{sec:methodology}

\subsection{One-Stage Policy}
The policy jointly selects the gate, the qubit indices, and the rotation angle in a single decision step per layer. Discrete components (gate and indices) are modeled via categorical distributions, while angles are produced by Gaussian heads and mapped to \([{-}\pi,\pi]\). To interface Proximal Policy Optimization (PPO) with the mixed discrete--continuous action, we parameterize a categorical distribution over the gate set and qubit indices together with a continuous distribution over angles; for implementations that share a common head, the unit interval can be partitioned into fixed bins that index gate categories, while a separate scalar is mapped to \([{-}\pi,\pi]\) via \(\theta=2\pi\tilde{\theta}-\pi\). We train with PPO \cite{ppo} using the clipped surrogate objective and standard entropy regularization. Algorithm~\ref{fig:one-stage-workflow} summarizes the one-stage workflow.

\subsection{Two-Stage Strategy}
The agent first proposes a discrete circuit topology by selecting gate types and placements without final angles. Afterwards, the rotation angles are optimized with Adam \cite{adam} using analytic gradients via the parameter-shift rule \cite{shift,shiftorigin}. We trigger the second stage when either half of the depth budget (\(L/2\)) is reached without satisfying the self-fidelity error (SFE) threshold, or at episode end otherwise. This decomposition isolates discrete structure selection from continuous refinement; Figure~\ref{fig:two-stage-workflow} illustrates the pipeline.

\subsection{Action Sampling and Hybrid Integration}
For an observation \(x_t\), the policy samples \(g_t\sim \mathrm{Cat}(\pi_g(x_t))\), \(q_t\sim \mathrm{Cat}(\pi_q(x_t))\), and \(\theta_t\sim \mathcal{N}(\mu_\theta(x_t), \sigma_\theta^2(x_t))\). The joint action is \(a_t=(g_t,q_t,\theta_t)\) with log-probability \(\log \pi(a_t\mid x_t)=\log\pi_g+\log\pi_q+\log\pi_\theta\). This explicit factorization clarifies how the next action is selected under a given policy and ensures that gradients propagate through both the discrete and continuous branches. Gymnasium provides episode bookkeeping and reward accumulation; PennyLane constructs the circuit specified by \(a_t\) and returns the updated statevector, which is then re-encoded as \(x_{t+1}\).

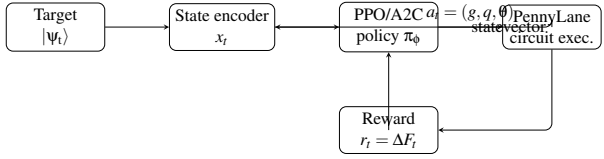
\begin{figure}[t]
    \centering
    \resizebox{0.95\columnwidth}{!}{%
    \begin{tikzpicture}[node distance=1.15cm,>=stealth,rounded corners]
        \tikzstyle{block} = [draw, rectangle, minimum width=2.0cm, minimum height=0.8cm, align=center]
        \node[block] (target) {Target\\$\lvert\psi_\mathrm{t}\rangle$};
        \node[block, right=1.3cm of target] (encoder) {State encoder\\$x_t$};
        \node[block, right=1.3cm of encoder] (policy) {PPO/A2C\\policy $\pi_\phi$};
        \node[block, right=1.3cm of policy] (pl) {PennyLane\\circuit exec.};
        \node[block, below=1.1cm of policy] (reward) {Reward\\$r_t=\Delta F_t$};
        \draw[->] (target) -- (encoder);
        \draw[->] (encoder) -- (policy);
        \draw[->] (policy) -- node[above]{\(a_t=(g,q,\theta)\)} (pl);
        \draw[->] (pl) |- (reward);
        \draw[->] (reward) -| (policy);
        \draw[->] (pl) |- node[left]{statevector} (encoder);
    \end{tikzpicture}%
    }
    \caption{Hybrid loop: Gymnasium manages the RL interface, PennyLane executes the quantum circuit, and the policy samples discrete gates, qubits, and continuous angles.}
    \label{fig:hybrid-pipeline}
\end{figure}

\subsection{Optimization Details}
PPO uses minibatch updates over collected rollouts with generalized advantage estimation. We grid-search learning rates and SFE thresholds; the most robust settings are \(\alpha\!\approx\!5\times 10^{-4}\) with an SFE of 0.01 for the one-stage policy, and \(\alpha\!\approx\!10^{-4}\) for the two-stage variant. A2C is included as a baseline but does not learn competitive policies in our environment. We report step counts and repetitions per experiment in Section~\ref{sec:results}.

\subsection{Experimental Setup}
The RL loop is implemented in Gymnasium and trained with PPO from \texttt{stable-baselines3}; actions proposed by the policy (gate type, qubit indices, angle) are consumed by the Gymnasium environment, which invokes PennyLane to construct and execute circuits on \texttt{default.qubit}. In the two-stage setting, PennyLane provides parameter-shift gradients that Adam uses to refine rotation angles, while PPO updates the policy parameters on the classical host. Tasks span \(n\in\{2,3,4,5,10\}\) qubits and complexity \(\lambda\in\{1,2,3,4,5\}\) with \(L=2\lambda\). Metrics include success rate under SFE, reconstructed circuit depth (RCD), and fidelity trajectories. For reproducibility, we average over multiple runs per configuration and report counts alongside results. We rely on Gymnasium for the reinforcement-learning interface and episode management and on PennyLane for hybrid quantum--classical execution and gradient evaluation \cite{gymnasium,bergholm2022pennylaneautomaticdifferentiationhybrid}.

\subsection{Rationale}
The one-stage policy aligns exploration and credit assignment across discrete and continuous choices but must cope with a larger action space. The two-stage strategy can refine angles efficiently given a reasonable discrete scaffold, but it cannot fix suboptimal discrete choices; empirically, it offers marginal accuracy gains at approximately three times the runtime.

\begin{algorithm}[!h]
 \caption{One-Stage Quantum Circuit Construction under PPO} \label{fig:one-stage-workflow}
 Initialize empty circuit $\mathcal{C}$ and policy parameters $\phi$\;
 \While{circuit not complete and step $< L$}{
   Select next \textbf{action} according to policy $\pi_\theta$\;
   \eIf{action = CNOT}{
      Choose qubit pair $(q_i, q_j)$\;
      Append CNOT$(q_i, q_j)$ to $\mathcal{C}$\;
   }{
      \If{action = Rotation-Gate}{
         Choose rotation gate type $R_\alpha(\theta)$ and target qubit $q_i$\;
         Initialize rotation parameter $\theta$\;
         \Repeat{parameter convergence}{
            Evaluate circuit performance (loss $\mathcal{L}$)\;
            Update $\theta \leftarrow \text{Adam}(\theta, \nabla_\theta \mathcal{L})$\;
         }
         Append $R_\alpha(\theta)$ to $\mathcal{C}$\;
      }
   }
 }
 Return optimized circuit $\mathcal{C}$\;
\end{algorithm}

\begin{algorithm}[!h]
 \caption{Two-Stage Quantum Circuit Construction and Parameter Optimization}
  \label{fig:two-stage-workflow}

 Initialize empty circuit $\mathcal{C}$\;
 \While{$(1 - \text{Fidelity}(\mathcal{C})) > \text{SFE}$}{
   Select next \textbf{discrete} action according to policy $\pi_\theta$\;
   Apply selected action (add or modify gate in $\mathcal{C}$)\;

   \If{step $=$ $L/2$ \textbf{or} $L$}{
      \tcp{Intermediate or final parameter optimization}
      \For{$t = 1$ \KwTo $T_{\text{opt}}$}{
         Evaluate loss $\mathcal{L}$ and compute gradients via the parameter-shift rule\;
         Update parameters using Adam: $\theta \leftarrow \text{Adam}(\theta, \nabla_\theta \mathcal{L})$\;
         \If{converged or budget exhausted}{
            \textbf{break}\;
         }
      }
   }
 }
 \Return{$\mathcal{C}$}
\end{algorithm}

\section{Results}
\label{sec:results}

We report results along four axes: hyperparameter selection, overall DQCS performance as a function of target complexity, state reconstruction on a representative benchmark set, and runtime characteristics of the two training strategies. Throughout, we also track reconstructed circuit depth (RCD), defined as the synthesized circuit depth expressed as a percentage relative to a reference depth associated with the target specification (lower is better; 100\% means matching the reference depth).

\subsection{Hyperparameter Selection}
We first established stable hyperparameter regimes using small targets (\(n=2\)) and moderate complexity (\(\lambda=2\), thus \(L=4\)). Across three runs per configuration (approximately three million steps per run), A2C did not yield learning beyond chance under multiple learning rates (\(\alpha\in\{10^{-2},10^{-3},10^{-4},10^{-5}\}\)). By contrast, PPO consistently improved the reconstructed depth and fidelity. The most robust settings were \(\alpha\approx5\times10^{-4}\) for the one-stage agent and \(\alpha\approx10^{-4}\) for the two-stage variant, with a self-fidelity-error threshold of \(\varepsilon=0.01\). We adopt these values in subsequent experiments unless noted. In the two-stage pipeline, Adam-based angle refinement converges within roughly 300 steps.

\subsection{DQCS Landscape versus Complexity}
\begin{figure}[t]
  \centering
  \begin{subfigure}[t]{0.99\columnwidth}
    \centering
    \includegraphics[width=\linewidth]{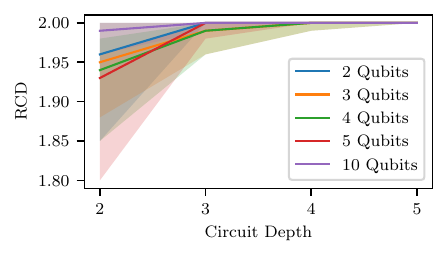}
    \caption{One-stage DQCS performance versus \(\lambda\).}
      \vspace{1em}  
  \end{subfigure}
  \begin{subfigure}[t]{0.99\columnwidth}
    \centering
    \includegraphics[width=\linewidth]{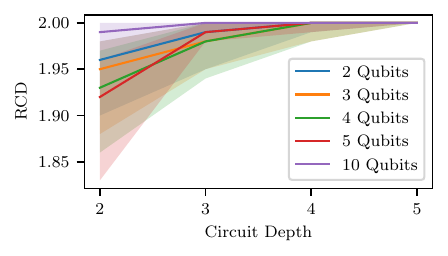}
    \caption{Two-stage DQCS performance versus \(\lambda\).}
    \vspace{1em}
  \end{subfigure}
  \caption{DQCS landscape across target complexity \(\lambda\) and qubit counts. The two-stage strategy occasionally improves fidelity but requires around three times the runtime. Mean values in solid lines, 95\% confidence intervals in shaded regions.}
  \label{fig:dqcs-landscape}
\end{figure}

We then assessed synthesis performance across target complexity by varying \(\lambda\in\{1,2,3,4,5\}\) for \(n\in\{2,3,4,5,10\}\). For systems with two to five qubits, both one-stage and two-stage agents achieve reliable preparation for \(\lambda\leq 2\) and exhibit diminishing returns thereafter, with success saturating around \(\lambda\leq 3\!\text{--}\!4\). At \(\lambda=5\), learning becomes inconsistent and sensitive to random seeds. For ten-qubit targets, neither approach achieves consistent success even at \(\lambda=2\), indicating a pronounced scalability boundary under our formulation. The two-stage approach occasionally yields slight fidelity gains near the SFE threshold but does not materially shift the boundary in either \(\lambda\) or \(n\) (Figure~\ref{fig:dqcs-landscape}).

\subsection{State Reconstruction Benchmarks}
To validate the environment and the learned policies, we evaluated reconstruction of canonical states. For two-qubit basis states, the one-stage agent reaches success rates between 83\% and 99\% depending on the target and \(\lambda\), with the highest reliability observed for \(\ket{00}\) at the upper end of the tested \(\lambda\) range. For Bell states, success rates range from 61\% to 77\% with best performance at higher \(\lambda\) within the tested range, reflecting the additional difficulty due to entanglement requirements. These results confirm that, within the identified complexity limits, the learned policies and environment produce correct circuits for both separable and entangled targets.
\subsection{Constructed Circuits and Numerical Example}
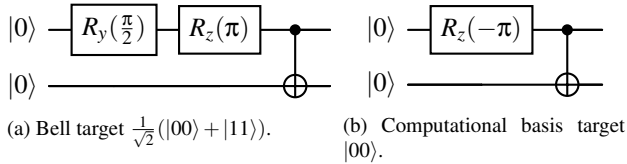
\begin{figure}[t]
    \centering
    \begin{subfigure}[t]{0.45\columnwidth}
        \centering
        \begin{quantikz}[column sep=0.3cm,row sep=0.25cm]
            \lstick{$\ket{0}$} & \gate{R_y(\frac{\pi}{2})} &\gate{R_z(\pi)} & \ctrl{1} & \qw \\
            \lstick{$\ket{0}$} & \qw & \qw & \targ{} & \qw
        \end{quantikz}
        \caption{Bell target \(\frac{1}{\sqrt{2}}(\ket{00}+\ket{11})\).}
    \end{subfigure}
    \hfill
    \begin{subfigure}[t]{0.45\columnwidth}
        \centering
        \begin{quantikz}[column sep=0.3cm,row sep=0.25cm]
            \lstick{$\ket{0}$} & \gate{R_z(-\pi)} & \ctrl{1} & \qw \\
            \lstick{$\ket{0}$} & \qw & \targ{} & \qw
        \end{quantikz}
        \caption{Computational basis target \(\ket{00}\).}
    \end{subfigure}
    \caption{Example circuits synthesized by the one-stage agent. Rotation gates align single-qubit phases; \(\mathrm{CNOT}\) gates propagate amplitude correlations.}
    \label{fig:circuit-examples}
\end{figure}

Reviewer requests for explicit gate usage are addressed in Figure~\ref{fig:circuit-examples}, which shows two agent-synthesized circuits together with their roles. The left panel prepares the Bell state \(\frac{1}{\sqrt{2}}(\ket{00}+\ket{11})\) with a single entangling layer (success rate 75\% at \(\lambda=7\)); the right panel shows a representative reconstruction of \(\ket{00}\) (success rate 99\% at \(\lambda=6\)). The rotation gates set local amplitudes and phases, while the \(\mathrm{CNOT}\) establishes entanglement or parity as needed. Both circuits were synthesized directly by the one-stage PPO agent and reach the SFE threshold \((1-F\leq 0.01)\).

\subsection{One-Stage versus Two-Stage and Runtime}
The two-stage strategy—reinforcement learning (RL) for discrete structure followed by Adam-based angle refinement with parameter-shift gradients—is intuitively appealing. Empirically, however, it offers only marginal accuracy benefits relative to the one-stage agent while incurring approximately three times the wall-clock training time due to repeated circuit evaluations for gradient-based optimization. Crucially, the two-stage approach cannot correct poor discrete gate placements; when the initial topology is suboptimal, angle refinement provides only limited recovery. Overall, the one-stage PPO agent is the more practical choice under a fixed compute budget. For completeness, we also verified that A2C failed to learn across all tested learning rates \(\alpha\in\{10^{-2},10^{-3},10^{-4},10^{-5}\}\).

\subsection{Classical Baseline and Literature Comparison}
To contrast with a purely classical optimizer, we trained a two-layer hardware-efficient ansatz with \(\{R_y,R_z,\mathrm{CNOT}\}\) blocks using Adam and parameter-shift gradients on ten random two-qubit targets with \(L=4\). After 300 gradient steps per target, the mean fidelity was 0.31 (minimum \(1.8\times 10^{-4}\)), markedly below the PPO one-stage success rates on the same target class (between 0.83 and 0.99 for basis states and 0.61 to 0.77 for Bell states). This gap indicates that the learned policy leverages discrete gate placement beyond what angle-only optimization provides. Compared with discrete DQCS edit policies \cite{rw1,tom} and curriculum-driven architecture search \cite{rw2}, our mixed discrete--continuous policy reaches similar reconstruction quality on small systems while exposing the scalability limits that arise once continuous parameters are incorporated.

\section{Discussion and Limitations}
\label{sec:discussion}

\subsection{Continuous Parameters and Credit Assignment}
Extending discrete gate selection to the joint selection of gates and continuous angles expands the effective action space and complicates temporal credit assignment. The policy must discover an appropriate discrete structure and tune angles simultaneously under relatively sparse terminal signals. While PPO provides a stabilizing inductive bias for mixed discrete--continuous choices, it does not fundamentally change the underlying optimization landscape. This observation is consistent with the modest empirical benefits of adding a second optimization stage.

\subsection{Limited Gains of Two-Stage Optimization}
Two-stage training refines angles efficiently when the proposed discrete scaffold is reasonable, leveraging Adam with parameter-shift gradients. However, it cannot repair suboptimal gate placements. In our setting, the discrete topology chosen by the agent largely determines final quality; post hoc angle refinement therefore adds limited value while increasing runtime due to repeated circuit evaluations.

\subsection{Goal Attainment and Hybrid Integration}
The research goal was to learn parameterized circuits that reach high fidelity under a compact gate universe. Within \(\lambda\leq 3\!\text{--}\!4\) and up to five qubits, the one-stage PPO agent achieves this objective and produces explicit circuits for separable and entangled targets. The hybrid Gymnasium--PennyLane pipeline (Figure~\ref{fig:hybrid-pipeline}) made this feasible by exposing statevectors to the policy while delegating circuit execution and gradients to PennyLane. The classical variational baseline falls well short of PPO on the same targets, which indicates that coupling discrete topology learning with continuous angles is essential even in small systems.

\subsection{Scalability and Entanglement}
The empirical boundary (\(\lambda\leq 3\!\text{--}\!4\) for systems with two to five qubits; failure at ten qubits even for \(\lambda=2\)) suggests that exploration and optimization pressure dissipate as target complexity and horizon length increase. Entanglement requirements exacerbate this effect, which is reflected in lower success on Bell states relative to basis states. Improving scalability will likely require structured exploration (for example, curriculum learning \cite{rw2}), better initialization of angles and gate placements, or hierarchical decompositions that constrain search to promising substructures.

\subsection{Practical Implications}
For parameterized state preparation under tight training budgets, a one-stage PPO agent with carefully chosen hyperparameters (\(\alpha\approx5\times10^{-4}\), SFE=0.01) is preferable to a two-stage pipeline. When high-fidelity targets at larger \(\lambda\) or higher \(n\) are required, augmenting reinforcement learning with stronger priors or hybrid search appears necessary. We also observe a practical limitation of the gate set: Hadamard operations are implemented by a sequence of five gates, which increases depth; adding \(H\) to \(\mathcal{G}\) or improving the decomposition could reduce overhead.

\subsection{Threats to Validity}
Results may depend on random seeds, specific target generation procedures, and backend execution noise models. We mitigate variance by averaging over multiple runs per configuration, but larger-scale replications and alternative target sets would strengthen external validity. Early stopping based purely on SFE may prematurely terminate borderline cases; more informed criteria (for example, gradient-based tangent checks) could help.

\section{Conclusion}
\label{sec:conclusion}

We introduced a reinforcement-learning environment for parameterized quantum state preparation that integrates continuous rotation angles into a DQCS-style setting and compared two training strategies: a one-stage policy that jointly selects gates and angles, and a two-stage approach that refines angles with Adam using parameter-shift gradients. Across systems with two to ten qubits and depth budgets with \(\lambda\in\{1,\dots,5\}\), A2C failed to learn, whereas PPO yielded stable improvements with clear hyperparameter regimes (one-stage: \(\alpha\approx5\times10^{-4}\) with SFE of 0.01; two-stage: \(\alpha\approx10^{-4}\)). Both strategies reconstruct basis and Bell states reliably within modest complexity, but scalability saturates around \(\lambda\leq 3\!\text{--}\!4\) and collapses at ten qubits even for \(\lambda=2\).

The two-stage method provides at most marginal fidelity gains at around three times the runtime and cannot recover from poor gate placements, making the one-stage PPO policy the pragmatic choice under compute constraints. A classical variational baseline with the same gate universe trails PPO markedly, underscoring the value of learning discrete topology jointly with continuous angles. These findings quantify the trade-offs that arise when extending discrete DQCS to continuous parameters and establish a baseline for future work.

Future directions include curriculum learning to stage target difficulty, improved parameter and topology initialization, and hybrid search methods that combine reinforcement learning with analytical synthesis or local heuristics. Extending the gate set (for example, including \(H\)) and refining early-stopping criteria beyond SFE (for example, tangent-based checks) are practical steps to reduce depth overhead and avoid premature termination. We believe these avenues are essential to push the scalability boundary and to make parameterized state preparation practical for larger systems.

\section*{Acknowledgments}
This paper was partially funded by the German Federal Ministry of Education and Research through the funding program ``Quantum Computing User Network'' (contract number 13N16196).
This research is part of the Munich Quantum Valley,
which is supported by the Bavarian state government
with funds from the Hightech Agenda Bayern Plus.

{
    \small
    \bibliographystyle{ieeenat_fullname}
    \bibliography{bibliography}
}

\end{document}